\documentclass{article}

\usepackage{microtype}
\usepackage{graphicx}
\usepackage{subcaption}
\usepackage{booktabs} 

\usepackage[colorlinks=true, allcolors=blue]{hyperref}

\usepackage[preprint]{icml2026}

\usepackage{amsmath}
\usepackage{amssymb}
\usepackage{mathtools}
\usepackage{amsthm}
\usepackage{placeins}   

\usepackage[capitalize,noabbrev]{cleveref}

\theoremstyle{plain}

\theoremstyle{definition}

\theoremstyle{remark}

\usepackage[textsize=tiny]{todonotes}

\usepackage{multirow} 
\usepackage{rotating}
\usepackage{physics}
\usepackage{enumitem,xspace}
\usepackage{makecell}
\setlist[enumerate]{leftmargin=*, topsep=0pt,itemsep=-1ex,partopsep=1ex,parsep=1ex}
\setlist[itemize]{leftmargin=*, topsep=0pt,itemsep=-1ex,partopsep=1ex,parsep=1ex}
\usepackage{subcaption}
\usepackage[svgnames]{xcolor}
\usepackage[most,listings]{tcolorbox}  
\usepackage{microtype}
\usepackage{caption}
\usepackage{courier} 
\tcbuselibrary{skins,breakable}

\newcommand{\cA}{\mathcal{A}}
\newcommand{\cS}{\mathcal{S}}

\newcommand{\E}{\mathbb{E}}

\newcommand{\think}{\texttt{<think>} \(\ldots\) \texttt{</think>}\xspace}
\newcommand{\answer}{\texttt{<answer>} \(\ldots\) \texttt{</answer>}\xspace}
\newcommand{\toolcall}{\texttt{<tool\_call>} \(\ldots\) \texttt{</tool\_call>}\xspace}
\newcommand{\st}{s^{\text{term}}}

\newcommand{\qwen}{\text{Qwen2.5-7B-Instruct}\xspace}

\newcommand{\eg}{\text{e.g.}}
\newcommand{\ie}{\text{i.e.}}

\definecolor{promptframe}{HTML}{2A6099} 
\definecolor{promptbg}{HTML}{EAF1F5}    
\definecolor{responseframe}{HTML}{B8860B} 
\definecolor{responsebg}{HTML}{FFF8DC}   

\newtcblisting{promptbox}[1][]{
  colback=promptbg,
  colframe=promptframe,
  fonttitle=\bfseries,
  title=#1,
  breakable,
  enhanced,
  attach boxed title to top left={yshift=-2mm, xshift=3mm},
  boxed title style={
    colback=promptframe,
    sharp corners,
    boxrule=0pt,
  },
  listing only, 
  listing options={
    basicstyle=\rmfamily\small\sloppy, 
    breaklines=true, 
    breakatwhitespace=true, 
    breakindent=0pt, 
    literate=
        {<think>}{{\bfseries <think>}}{7}
        {</think>}{{\bfseries </think>}}{8}
        {<tool_call>}{{\bfseries <tool\_call>}}{11}
        {</tool_call>}{{\bfseries </tool\_call>}}{12}
        {<tool_response>}{{\bfseries <tool\_response>}}{15}
        {</tool_response>}{{\bfseries </tool\_response>}}{16}
        {<answer>}{{\bfseries <answer>}}{8}
        {</answer>}{{\bfseries </answer>}}{9}
        {<verification>}{{\bfseries <verification>}}{14}
        {</verification>}{{\bfseries </verification>}}{15}
        {<verification_result>}{{\bfseries <verification\_result>}}{21}
        {</verification_result>}{{\bfseries </verification\_result>}}{22}
  }
}

\newtcblisting{responsebox}[1][]{
  colback=responsebg,
  colframe=responseframe,
  fonttitle=\bfseries,
  title=#1,
  breakable,
  enhanced,
  attach boxed title to top left={yshift=-2mm, xshift=3mm},
  boxed title style={
    colback=responseframe,
    sharp corners,
    boxrule=0pt,
  },
  listing only, 
  listing options={
    basicstyle=\rmfamily\small\sloppy, 
    breaklines=true, 
    breakatwhitespace=true, 
    breakindent=0pt, 
    literate=
        {<think>}{{\bfseries <think>}}{7}
        {</think>}{{\bfseries </think>}}{8}
        {<tool_call>}{{\bfseries <tool\_call>}}{11}
        {</tool_call>}{{\bfseries </tool\_call>}}{12}
        {<tool_response>}{{\bfseries <tool\_response>}}{15}
        {</tool_response>}{{\bfseries </tool\_response>}}{16}
        {<answer>}{{\bfseries <answer>}}{8}
        {</answer>}{{\bfseries </answer>}}{9}
        {<verification>}{{\bfseries <verification>}}{14}
        {</verification>}{{\bfseries </verification>}}{15}
        {<verification_result>}{{\bfseries <verification\_result>}}{21}
        {</verification_result>}{{\bfseries </verification\_result>}}{22}
  }
}

\icmltitlerunning{Rethinking the Design of Reinforcement Learning-Based Deep Research Agents}
\allowdisplaybreaks
\begin{document}

\twocolumn[
  \icmltitle{Rethinking the Design of Reinforcement Learning-Based Deep Research Agents}



  \icmlsetsymbol{equal}{*}

  \begin{icmlauthorlist}
    \icmlauthor{Yi Wan}{equal,Pokee AI}
    \icmlauthor{Jiuqi Wang}{equal,Pokee AI}
    \icmlauthor{Liam Li}{Pokee AI}
    \icmlauthor{Jinsong Liu}{Pokee AI}
    \icmlauthor{Ruihao Zhu}{Pokee AI}
    \icmlauthor{Zheqing Zhu}{Pokee AI}
  \end{icmlauthorlist}

  \icmlaffiliation{Pokee AI}{Pokee AI}

  \icmlcorrespondingauthor{Yi Wan}{yi.wan@pokee.ai}

  \icmlkeywords{Machine Learning, ICML}

  \vskip 0.3in
]



\printAffiliationsAndNotice{\icmlEqualContribution}

\begin{abstract}
Large language models (LLMs) augmented with external tools are increasingly deployed as deep research agents that gather, reason over, and synthesize web information to answer complex queries. Although recent open-source systems achieve strong empirical performance via reinforcement learning from web interactions, the impact of key design choices remains under-explored. We formalize deep research as reinforcement learning in an episodic finite Markov decision process and construct a competitive baseline agent grounded in this formulation. Building on this foundation, we systematically examine critical design decisions at both training and inference time and identify four factors that substantially improve performance: replacing rule-based rewards with AI feedback from an LLM judge, fine-tuning with the on-policy RLOO algorithm instead of the off-policy GRPO algorithm, filtering low-quality training samples, and employing an error-tolerant test-time rollout strategy. Together, these design choices yield a deep research agent that establishes state-of-the-art performance among 7B-scale agents when evaluated across ten widely used benchmarks.
\end{abstract}

\section{Introduction}
Large language models (LLMs) equipped with external tools have enabled a new class of systems --- deep research agents --- that can gather information, reason across heterogeneous sources, and synthesize evidence to answer complex, open-ended queries. 
Since the introduction of OpenAI’s Deep Research agent \cite{openai2025deepresearch}, both proprietary agents \cite{google_gemini_deep_research_overview_2025,perplexity_deep_research_2025,moonshot_kimi_researcher_2025} and a growing body of open-source agents have demonstrated strong empirical performance across diverse research-oriented benchmarks.

Among open-source efforts, many works treat deep research as a reinforcement learning (RL) problem \cite{jin2025search,shi2025search,zheng2025deepresearcher,mei20252,tian2025ego,wang2025stepsearch,dong2025tool,li2025websailor,chen2025research,shi2025deepdiver,wu2025webdancer,tao2025webleaper}. 
Here, the web is treated as an environment, and the agent, which is initialized from a pretrained LLM, learns a policy through interaction data to maximize the likelihood of producing correct answers. 
Most RL-based deep research agents adopt the ReAct framework \citep{yao2023react}, where tool calls and answer generation are conditioned on intermediate reasoning tokens, and the agent’s state consists of the whole interaction history. 
While these agents consistently achieve strong benchmark results, they are typically presented as end-to-end pipelines comprising many intertwined design choices, making it difficult to assess which components are essential to their success.

Only a limited number of works isolate critical design decisions in their agents. 
Notable works include those studying the benefit of incorporating supervised learning signals alongside RL \citep{zhang2025evolvesearch,dong2025tool}, the effectiveness of carefully engineered, rule-based reward functions that encode priors over answer quality, reasoning structure, and tool usage \citep{mei20252,shi2025search,wang2025stepsearch,dong2025tool}, and the advantage of using diversified and challenging training data \citep{song2025r1}.

In this work, we advance the understanding of critical design choices in deep research agents by conducting a controlled empirical study of key design choices in ReAct-based agents. 
Building on a rigorously formulated RL framework for deep research, we develop a potent base agent and perform targeted ablations to identify which components materially impact performance. 
Our investigation yields four main findings:

1) We show that using AI feedback as the training reward substantially outperforms the F1 score, a rule-based reward obtained by calculating word-level similarity between a predicted answer and the ground truth. 
F1 score has been adopted in the training of many existing deep research agents \citep{mei20252,shi2025search,wang2025stepsearch,dong2025tool,song2025r1,zhang2025evolvesearch,zheng2025deepresearcher}. 
Note that deep research requires evaluating free-form, long-form answers. 
LLM-based feedback provides much more accurate learning signals than F1 score and rule-based rewards in general. 
In addition, we found that format rewards, which are auxiliary rewards encouraging correct tool call and answer format and widely adopted in existing deep research agents \citep{zhang2025evolvesearch,mei20252,song2025r1,dong2025tool}, are unnecessary;

2) We demonstrate that REINFORCE Leave-One-Out (RLOO) \cite{kool2019reinforce} is significantly more sample-efficient than the prevalent Group Relative Policy Optimization (GRPO) \cite{shao2024deepseekmath} for fine-tuning deep research agents. 
We show that this advantage stems from RLOO’s on-policy nature rather than from removing the length normalization and advantage normalization biases in GRPO.

3) We identify training data curation as a critical factor.
Beyond including more challenging and diversified data, a technique also noted by \citet{song2025r1}, we show that filtering low-quality samples using AI feedback and adjusting difficulty levels based on the initial policy's pass@k metric provides stronger learning signals.

4) We find that an error-tolerant test-time rollout strategy further improves performance. 
Instead of terminating episodes upon encountering errors such as invalid tool calls or malformed outputs, allowing the agent to recover and continue its rollout at test time leads to an additional accuracy gain.

Together, these four design choices yield a deep research agent that establishes state-of-the-art performance among 7B-scale agents when evaluated across ten widely used benchmarks.

\section{Problem Formulation and Base Agent}\label{sec:formulation}
We formalize the deep research task as a reinforcement learning problem in which an agent interacts with a stochastic environment modeled as an episodic finite Markov decision process (MDP) $M=(\cS, \cA, \mathcal{R}, p, r, \st)$. 
Here, $\cS$, $\cA$, $\mathcal{R}$ denote the finite state, action, and reward spaces, respectively, $p:\cS\times \cA \to \Delta(\cS)$ is the state transition kernel, where $\Delta$ is the probability simplex, $r:\cS\times\cA\times\cS \to\Delta(\mathcal{R})$ is the (possibly random) reward function, and $\st \in \cS$ is an absorbing terminal state. 
Each episode consists of a sequence of agent-environment interactions. 
In each time period $t=0,1,\ldots, $ the agent observes the current state $S_t\in \cS$, selects an action $A_t \sim \pi(\cdot \mid S_t)$ following a policy $\pi : \cS \to \Delta(\cA)$, and then observes the next state $S_{t+1} \sim p(\cdot \mid S_t, A_t)$ along with a reward $R_{t+1} \sim r(S_t, A_t, S_{t+1})$.
By definition of $\st$, $p(\st\mid \st, a) =1 \text{, and } r(\st, a, \st) = 0 ~\forall a\in\cA$. 
To ensure the problem is well-defined, we assume that all policies reach the terminal state almost surely in a finite number of steps. 
It will soon be clear that this assumption holds in the deep research task. 
With this assumption, the expected cumulative reward $\E[\sum_{t=0}^{T-1} R_{t+1} | S_0 = s]~ \forall s \in \cS$ is well-defined. 
Here, $T$ is a random variable denoting the episode length; the expectation is w.r.t. the randomness of the transition kernel, the reward, and the policy (together with the induced $T$). Let $\Pi$ denote a set of representable policies.
The agent's goal is to search for a policy $\pi \in \Pi$ that maximizes the expected return with states sampled from a pre-defined initial state distribution $\mu\in \Delta(\cS)$. 
That is, $\max_{\pi \in \Pi} \sum_{s \in \cS} \mu(s) \mathbb{E}_\pi\qty[\sum_{t=0}^{T-1} R_{t+1} \mid S_0 = s]$.

To describe a deep research task, we can let the action space $\cA$ be the universe of tokens, \ie, the dictionary of the LLM. 
Denoting $\cS'$ as the set of all possible token sequences with length at most $H$, where $H$ is a predefined integer, the state space can be defined as $\cS: \cS' \cup \st$. 
Here, $\st$ is a nominal state marking the termination of episodes rather than a token sequence. 
The initial state distribution $\mu$ assigns a non-zero mass to every sequence of tokens encoding a predefined system prompt (\ie, a description of the agent’s task and the available information-seeking tools, such as web search and web reading) together with a user query, and assigns a zero mass to other sequences. 
The system can take actions (\ie, generate tokens) that trigger tool invocations or provide answers. 
We count each tool invocation or query answering as a turn, and the deep research task can run for at most $N$ turns, where $N$ is a predefined integer. 
Specifically, a turn is defined by two possible combinations of tags (and the tokens enclosed between them), namely \think followed by \answer, and
\think followed by \toolcall,
whose usage will be specified later. 
Note that this problem formulation follows the ReAct \cite{yao2023react} idea of synergizing intermediate auxiliary ``think'' tokens and the functional tokens (tool calls and the answer).
It is evident that a single turn typically spans multiple time periods in the MDP. 
Under this formulation, the state transitions and system dynamics can be described as follows:
\begin{enumerate}
\item If the action is not the End of Sequence (EOS) token and the length of the current state's token sequence is shorter than $H$, the MDP transitions to the next state, which is the concatenation of the current state's token sequence and the action token;

\item If the action is the EOS token, the current state's token sequence is shorter than $H$, and the turn counter (initialized to 0) has not reached $N$, the sequence of all action tokens in the current turn is decoded to a snippet of text. 

\begin{enumerate}
    \item If the text is wrapped by the \think \toolcall tags, the text between \toolcall is extracted to invoke a tool call. 
    Each tool call script must follow a specific format (\eg, specify required arguments). 
    If the specified tool exists and the format is correct, the tool call is executed, and the result is returned. 
    The MDP then transitions to the next state, which is the concatenation of the current state, the action, and a sequence of tokens encoding the tool call result (for simplicity, we view the tool call results as texts). 
    The turn counter is increased by 1.

     \item If the text is wrapped by the \think \answer tags, the text between \answer is extracted as the answer to the user query. The MDP goes to $\st$;

\end{enumerate}
\item In all other cases, the MDP goes directly to $\st$.
\end{enumerate}
All rewards are zero except at the end of the episode. 
If an episode finishes with an answer in between the two answer tags, a non-negative reward is given to the last transition based on the quality of the answer. 
Otherwise, the reward is 0. 

\noindent \textbf{Base Agent Design.} 
Our base agent adopts \qwen \cite{qwen2.5} as the policy $\pi$. 
\qwen is a powerful open-source LLM that has been instruction-tuned on a large corpus of human-generated data and has been commonly used in existing deep research agents. 
We use \qwen's associated tokenizer to convert between texts and tokens. 
Under this, $H=32k$ is \qwen's context length limit, and we set the maximum number of turns to be $T=10$. 
We provide two tools to our base agent for fetching web content.
\begin{itemize}
\item \textit{Web Searching Tool:} 
We use Serper \citep{serper2025} to facilitate web-based information retrieval. 
The tool accepts a list of string queries, runs searches via Google, and returns a structured set of URLs along with descriptive snippets for each query. 
This helps the agent to survey the information landscape and iteratively identify high-priority sources for deeper investigation;

\item \textit{Web Reading Tool:} 
This tool takes as input a list of \((\text{URL}, \text{query})\) pairs. 
For each pair, if the webpage contains information relevant to answering the query, it generates a brief answer based on the page’s content; otherwise, it returns a message indicating that the required information is not available. 
Internally, Jina Reader \citep{jina2025} retrieves and parses the webpage content for each URL, and Gemini-Flash-lite 2.5, a small LLM, answers the query solely based on that content. 
As a result, this combination produces more concise responses and helps prevent the context length from growing too rapidly.
\end{itemize}

Similar to prior work, we fine-tune the base policy to improve its ability to use tools. 
Following \citet{zheng2025deepresearcher}, we use three data sources Natural Questions (NQ) \citep{kwiatkowski2019natural}, TriviaQA (TQ) \citep{joshi2017triviaqa}, and 2WikiMultiHopQA (2Wiki) \citep{ho2020constructing} to serve as the training dataset. 
NQ and TQ consist of factual questions that often require only a single internet search to answer, and thus are relatively simple. 
Questions in 2Wiki are considered multi-hop, requiring the agent to combine multiple sources to reach the final answer. 
The training reward is chosen to differ from the test reward, since a powerful LLM assigns the test reward as a proxy for human evaluation. 
Training with human feedback is extremely costly and inefficient. 
Instead, we start by using the rule-based F1 score, which is the harmonic mean of precision and recall between the generated answer text and the ground-truth text (see Section \ref{sec:metric} for a formal definition), as the reward signal. 
This reward signal is also used to train many other deep research agents \cite{mei20252,shi2025search,wang2025stepsearch,dong2025tool,song2025r1,zhang2025evolvesearch,zheng2025deepresearcher}. 
The learning algorithm is a direct MDP extension of the GRPO algorithm, originally proposed under contextual bandits. 
See Section \ref{sec:grpo} for more details of this algorithm.  
Other details of the experiment are provided in Section \ref{sec:base_agent_fine_tune}.

\noindent \textbf{Testing Datasets.} 
To evaluate the base agent, we construct a test set of (question, ground-truth) pairs drawn from widely used deep research benchmarks. 
In addition to TQ, NQ, and 2Wiki, we include GAIA \cite{mialon2023gaia}, BrowseComp (BC) \cite{wei2025browsecomp}, Human’s Last Exam (HLE) \cite{phan2025humanity}, PopQA (POP), MuSiQue (MUS) \citep{trivedi2022musique}, Bamboogle (BAM) \citep{press2022measuring}, and HotpotQA (HOT) \cite{yang2018hotpotqa}. 
All questions, except those in TQ, NQ, and PopQA, are multi-hop.
Among them, GAIA, HLE and BC are more recent and significantly more challenging. 
A detailed description of each benchmark is provided in Appendix~\ref{sec:benchmarks}.

We observe that several benchmarks contain a nontrivial fraction of low-quality or incorrect $(\text{question}, \text{list of reference answers})$ pairs. 
For this, we submit all pairs from the seven affected QA benchmarks to Gemini-2.5-Pro, a state-of-the-art proprietary LLM, and use its judgments to remove low-quality pairs. 
Prompt for data cleaning and examples of low-quality question-answer pairs can be found in Section~\ref{sec:cleaning_prompt}. 
We then construct the final test set by randomly sampling 125 questions from each benchmark, except (1) GAIA contains only 103 text-only questions, all of which are used; and (2) after cleaning, PopQA, MuSiQue, and Bamboogle contain only 124, 116, and 83 questions, respectively, and we therefore include all remaining questions from these benchmarks. 
In total, the resulting test set comprises 1,176 questions. 

For each test question, we run four independent trials. 
If a run produces an answer, we submit the question, the generated answer, and the corresponding ground-truth answer to Gemini-2.5-Flash, a powerful, proprietary LLM, which evaluates the predicted answer's correctness and outputs a binary reward. 
Here, the LLM serves as a proxy for a human judge. 
Since all questions in the test set require only short answers and their ground-truth answers are provided, the evaluation task is unambiguous and straightforward; we therefore expect the LLM’s judgments to match those of human evaluators closely. 
To validate this assumption, we manually inspected 100 randomly sampled judgments and found strong agreement. 
If a run fails to produce an answer, the reward is set to zero.

\vspace{2mm}
\noindent \textbf{Results and Comparisons with Other Agents.}  
To demonstrate that the base agent provides a strong foundation for subsequent analysis, we compare its performance against several recent open-source deep research agents of the same scale (7B parameters). 
These include agents trained in simulated or static web environments, such as R1-Searcher \citep{song2025r1}, Search-R1 \citep{jin2025search}, and ZeroSearch \cite{sun2025zerosearch}, as well as agents trained with access to the live web, namely DeepResearcher \citep{zheng2025deepresearcher}, ASearcher \citep{gao2025turnsunlockinglonghorizonagentic}, and WebSailor \citep{li2025websailor}. 
We note that several other ReAct-based deep research agents evaluated at the same model scale have not released their 7B checkpoints \citep{zhang2025evolvesearch,shi2025deepdiver,wu2025webdancer,tao2025webshaper,team2025mirothinker}. Additionally, for a fair comparison, we exclude an agent trained with substantially longer context lengths \citep{liu2025webexplorer}. 
We are also aware of additional work trained in simulated or static web environments \citep{wang2025stepsearch,chen2025research,shi2025search}. 
However, as we demonstrate below, agents trained with static or simulated web data consistently underperform those trained with live web access. 
Consequently, omitting these agents in the comparison does not weaken our claim that the base agent is competitive among the prior 7B-scale deep research agents.

 \begin{table*}[ht]

\centering
\scriptsize{

\begin{tabular}{llccccccccccc}
\toprule
 &  & \multicolumn{3}{c}{\textbf{In-distribution}}
 & \multicolumn{7}{c}{\textbf{Out-of-distribution}}
 & \\
\cmidrule(lr){3-5}\cmidrule(lr){6-12}
\textbf{Type} & \textbf{Method}
& \textbf{2WIKI} & \textbf{TQ} & \textbf{NQ}
& \textbf{BAM} & \textbf{POP} & \textbf{MUS} & \textbf{HOT}
& \textbf{HLE} & \textbf{GAIA} & \textbf{BC}
& \textbf{AVG} \\
\midrule

\multirow[c]{3}{*}{\makecell{\textbf{Static \&}\\\textbf{simulated}\\\textbf{web}}}
& R1-Searcher & 61.6 & 65.0 & 66.2 & 62.4 & 65.1 & 51.5 & 62.6 & 4.13 & 4.89 & 0.80 & 40.78 \\
& Search-R1   & 78.4 & 74.2 & 79.2 & 75.3 & 77.2 & 61.0 & 72.8 & 11.10 & 18.69 & 0.60 & 50.87 \\
& ZeroSearch  & 17.6 & 31.4 & 30.0 & 53.9 & 39.7 & 11.4 & 13.8 & 6.96 & 8.37 & 0.40 & 18.76 \\

\midrule
\multirow[c]{5}{*}{\makecell{\textbf{Live}\\\textbf{web}}}
& ASearcher        & 84.4 & 84.6 & 87.2 & 74.4 & 81.9 & 64.9 & 84.8 & 11.40 & 16.91 & 2.61 & 57.57 \\
& DeepResearcher   & 85.40 & 79.80 & 89.60 & 78.31 & 81.05 & 62.78 & 79.80 & 10.22 & 20.63 & 2.20 & 56.64 \\

& WebSailor & 88.8 & \textbf{92.8} & 97.6 & 86.8 & \textbf{87.9} & 69.0 & \textbf{92.8} & 12.8 & 34.0 & 5.6 & 66.8 \\
& Our base agent  & \textbf{92.0} & 82.7 & 88.3
                   & 84.3 & 83.6 & 67.0 & 80.3
                   & 9.6 & 26.2 & 1.87 & 61.40 \\
& Our best agent & 90.8 & 92.6 & \textbf{97.8} & \textbf{92.8} & 86.3 & \textbf{81.0} & 92.0 & \textbf{17.6} & \textbf{49.2} & \textbf{6.2} & \textbf{71.07}\\
\bottomrule
\end{tabular}

}
\caption{Performance comparison of the considered deep research agents across widely used deep research benchmarks. Each entry reports the average evaluation reward (multiplied by 100) achieved by the corresponding agent on the corresponding benchmark. AVG denotes the average number of correctly answered questions over the entire test set. We report mean correct answer rates across four independent evaluation runs. Boldface indicates the highest mean performance. Full results, including standard errors, are provided in Table~\ref{tab:qa_comparison_updated}.}
\label{tab:compare_with_other_systems}
\end{table*}

As shown in Table~\ref{tab:compare_with_other_systems}, incorporating live web search during training substantially improves overall performance, highlighting the importance of live web interactions. 
Compared with other baselines, our base agent outperforms ASearcher and DeepResearcher on most benchmarks and underperforms WebSailor on most benchmarks.
Due to the complexity of deep research agents, many design choices may affect performance. 
While identifying which specific design choices account for the performance differences between our base agent and prior agents is beyond the scope of this work, we highlight several factors that are likely contributors. (1) ASearcher adopts an asynchronous training framework, which introduces additional off-policiness into the learning algorithm. 
(2) DeepResearcher was trained for substantially fewer steps (34 steps) than ours (320 steps). 
(3) WebSailor is trained on a dataset composed of carefully synthesized questions, which may drive better learning behavior. 
Overall, our base agent is competitive among 7B agents, laying a strong foundation for the subsequent analysis.

\section{Important Design Choices to Improve the Base Agent} \label{sec:reward}
Throughout our exploration, we find that several design choices can further boost our agent's performance. 
In this section, we introduce them, including reward design, training algorithm, data curation, and error-tolerant test-time rollout, by additionally conducting a sequence of experiments. 
Each experiment uses the same setting as its predecessor except for one change. Unless specified, all experiments are performed for 3 independent runs.

\subsection{AI Feedback as Training Rewards}
\begin{figure}[!htbp]
\centering\includegraphics[width=0.45\textwidth]{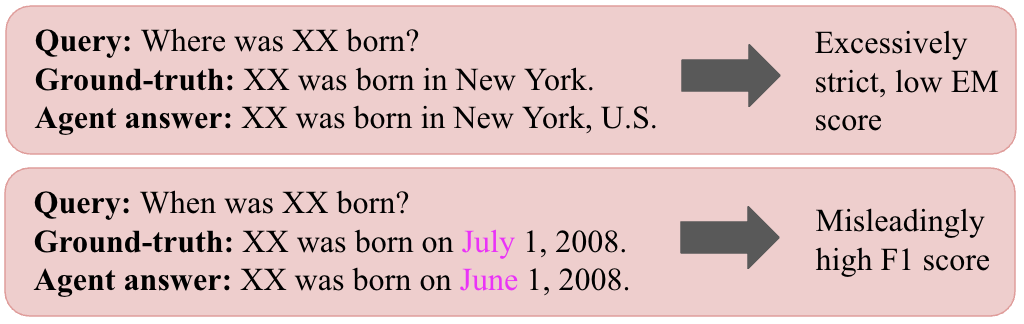}
    \caption{Illustration of the inadequacies of rule-based rewards.}
    \label{fig:mbe_advantages}
\end{figure}

An essential piece in RL fine-tuning is the training reward signal. 
Unlike in math and coding problems \citep{shao2024deepseekmath,chen2021evaluating}, it is non-trivial to verify the correctness of deep research's open-ended, free-form textual outputs. 
One approach adopted by a few works is to assign a positive reward only when the predicted answer matches a ground-truth answer. 
This exact-match (EM) approach suffers from excessive strictness, penalizing semantically correct answers that are not identical to the ground truth. 
The more commonly used training reward in existing works and our base agent is the F1 score. 
Despite its flexibility, it may produce misleadingly high scores for factually incorrect answers that share substantial token overlap with the ground truth. 
We use examples in Figure \ref{fig:mbe_advantages} to illustrate. 
Additional failure modes observed in our experiments are described in Section~\ref{sec:f1-limitations}.

\begin{figure*}[htbp]
    \centering
    \includegraphics[width=\linewidth]{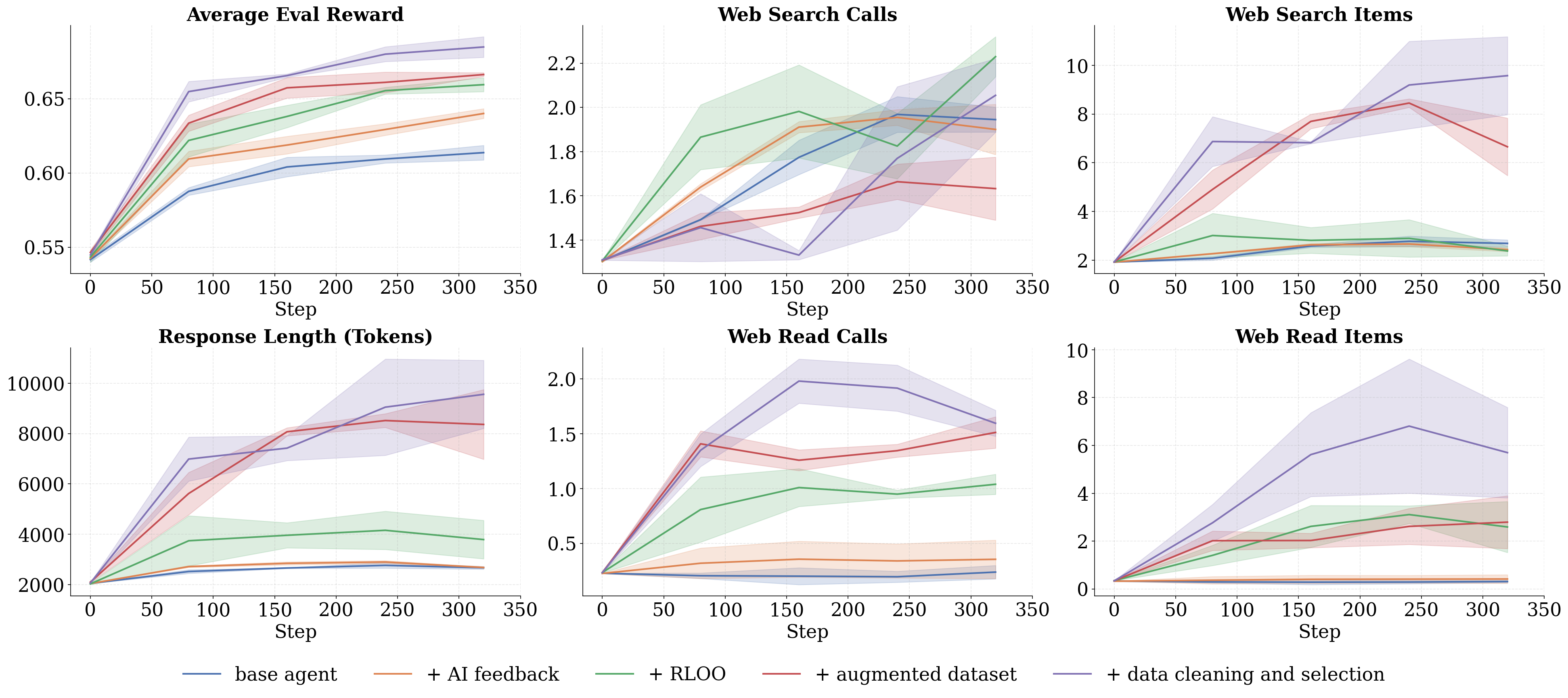}

    \caption{Dynamics of evaluation reward, response length, the number of tool calls and the number of query items involved in tool calls as training goes by. Each point in a curve is an average of three independent runs. The shading region indicates the standard error.}
    \label{fig:tool_call_stats}
\end{figure*}

To overcome the limitations of rule-based rewards, we adopt an AI-feedback approach that has been widely applied across domains \citep{gu2025judge}. 
In the context of deep research agents, \citet{liu2025webexplorer} similarly use an LLM --- DeepSeek-V3 \citep{ds2025deepseek} --- to assess the correctness of agent-generated answers. 
In this work, we demonstrate that reward signals produced by an LLM substantially less powerful than the evaluation model can yield much better performance than F1-score-based rewards (orange curves vs.\ blue curves in Figure~\ref{fig:tool_call_stats}). 
Specifically, training rewards are generated by Gemini-2.5-Flash-Lite, which is one-fifth the cost of the evaluation LLM and serves as a proxy for human judgment. 
Despite this significant cost difference, the two models exhibit high agreement (97.62\%). 
See Section~\ref{sec:ai-feedback-validation} for details of the disagreed cases. 
We further evaluate an even lower-cost model, GPT-5-Nano (half the price for input tokens and the same price for output tokens), and observe no clear performance difference (Figure~\ref{fig:mbe}). 
Taken together, these findings suggest that assessing semantic equivalence between predicted answers and ground-truth responses appears to be a simple task, for which inexpensive LLMs such as Gemini-2.5-Flash-Lite and GPT-5-Nano are sufficient. 

We also tested the widely used format reward, which assigns a small positive reward at the end of the episode when the episode ends with an answer and the answer is judged incorrect. However, our results (Figure \ref{fig:format reward}) show that this has almost no impact on performance.

\subsection{RLOO as Training Algorithm}

Next, we show that replacing the off-policy GRPO algorithm with the on-policy RLOO algorithm \cite{kool2019reinforce} can achieve much higher performance. 
In addition to this on-/off-policy difference, GRPO also introduces multiple sources of biases, while RLOO does not.

\noindent\textbf{Biases of GRPO.} \cite{liu2025understanding} observe that GRPO suffers from two kinds of bias, causing the gradient of \(L_{\text{GRPO}}\) to deviate from an unbiased policy gradient estimate. 
\begin{enumerate}
    \item GRPO normalizes returns by their standard deviation. 
    This implicitly assigns higher weight to tasks with narrower return distributions, which are often either too easy or too difficult;
    \item It normalizes the loss function by episode length, which reduces the magnitude of updates for longer trajectories.
\end{enumerate}
To address the above issues, \cite{liu2025understanding} propose Dr. GRPO. Nevertheless, we identify two additional biases in (Dr.) GRPO.
\begin{enumerate}
    \item (Dr.) GRPO samples episodes using an outdated policy, making it an off-policy algorithm. 
    This necessitates the use of an importance sampling ratio in the updates, which can result in high variance. 
    Therefore, a clipping operation is deployed to improve stability, which introduces bias, and
    \item (Dr.) GRPO normalizes the loss function directly by the sample size (see Appendix \ref{sec:grpo}), which introduces a multiplicative bias.
\end{enumerate}
The first bias appears to be inherent to off-policy algorithms, as otherwise the variance of the policy gradient estimate can become extremely high. 
The second bias, in contrast, can be easily absorbed by using a slightly higher learning rate and thus has no impact on optimization.

\noindent\textbf{RLOO.} 
RLOO introduces a slight modification to the classic REINFORCE algorithm \citep{williams1992simple} that reduces variance at the cost of sampling multiple trajectories. 
Although developed initially for general probabilistic models, RLOO has since been applied to contextual bandits \citep{ahmadian2024back}. 
Here, we present how to adapt it for MDPs with a detailed derivation deferred to Appendix~\ref{section: RLOO derivation}.

At each iteration, RLOO samples $m$ i.i.d. initial states $S_0^1, \dots, S_0^m$, 
where $n$ i.i.d. episodes are sampled with the current policy $\pi_\theta$ rather than an older policy starting from each initial state, resulting in $m\cdot n$ trajectories $S_0^i, A_0^{i,j}, R_1^{i,j}, \dots, S_{T^{i,j}}^{i,j}, \forall i = 1, \dots, m, \forall j = 1,\dots,n$.
Then, RLOO updates the policy parameters $\theta$ by ascending the gradient of the following objective:
\begin{equation}
\begin{aligned}
    &L_{\text{RLOO}}(\theta)\\
    \doteq&
    \frac{1}{m}
    \sum_{i=1}^m
    \frac{1}{n-1} \sum_{j=1}^n 
    \sum_{t=0}^{T^{i,j} - 1} \hat{A}^{i,j} \log \pi_\theta\qty(A^{i,j}_t \mid S^{i,j}_t),
    \label{eq:RLOO}
\end{aligned}
\end{equation}
where $\hat{A}^{i,j} \doteq G^{i,j} - \bar{G}^i$, $\bar{G}^i \doteq \frac{1}{n} \sum_{k=1}^n G^{i,k}$, and
$G^{i,j} \doteq \sum_{t=0}^{T^{i,j} - 1} R_{t+1}^{i,j}$.
It can be shown that \(\nabla_\theta L_{\text{RLOO}}\) is an unbiased estimator of the policy gradient of the expected cumulative reward \(\mathbb{E}_{\pi_\theta}\!\left[\sum_{t=0}^{T-1} R_{t+1}\right]\) (see Appendix~\ref{section: RLOO derivation}).

To evaluate RLOO, we replaced GRPO with RLOO in our agent trained using AI-generated feedback as rewards, while keeping all other hyperparameters unchanged. 
Comparing the RLOO's and GRPO's learning curves (green and orange curves in Figure \ref{fig:mbe_advantages}) reveals that RLOO achieves higher performance. 
One additional observation from this figure is that, as learning progresses, RLOO learns to generate more extended sequences and more web read calls than GRPO. 
One possible explanation for this behavior is that RLOO's on-policy nature allows it to deviate further from the initial policy with the same number of samples/updates. 
This might also be the reason why RLOO achieves a higher reward.

\begin{figure*}[htbp]
    \centering
    \begin{subfigure}{0.325\textwidth}
        \centering
        \includegraphics[width=\textwidth]{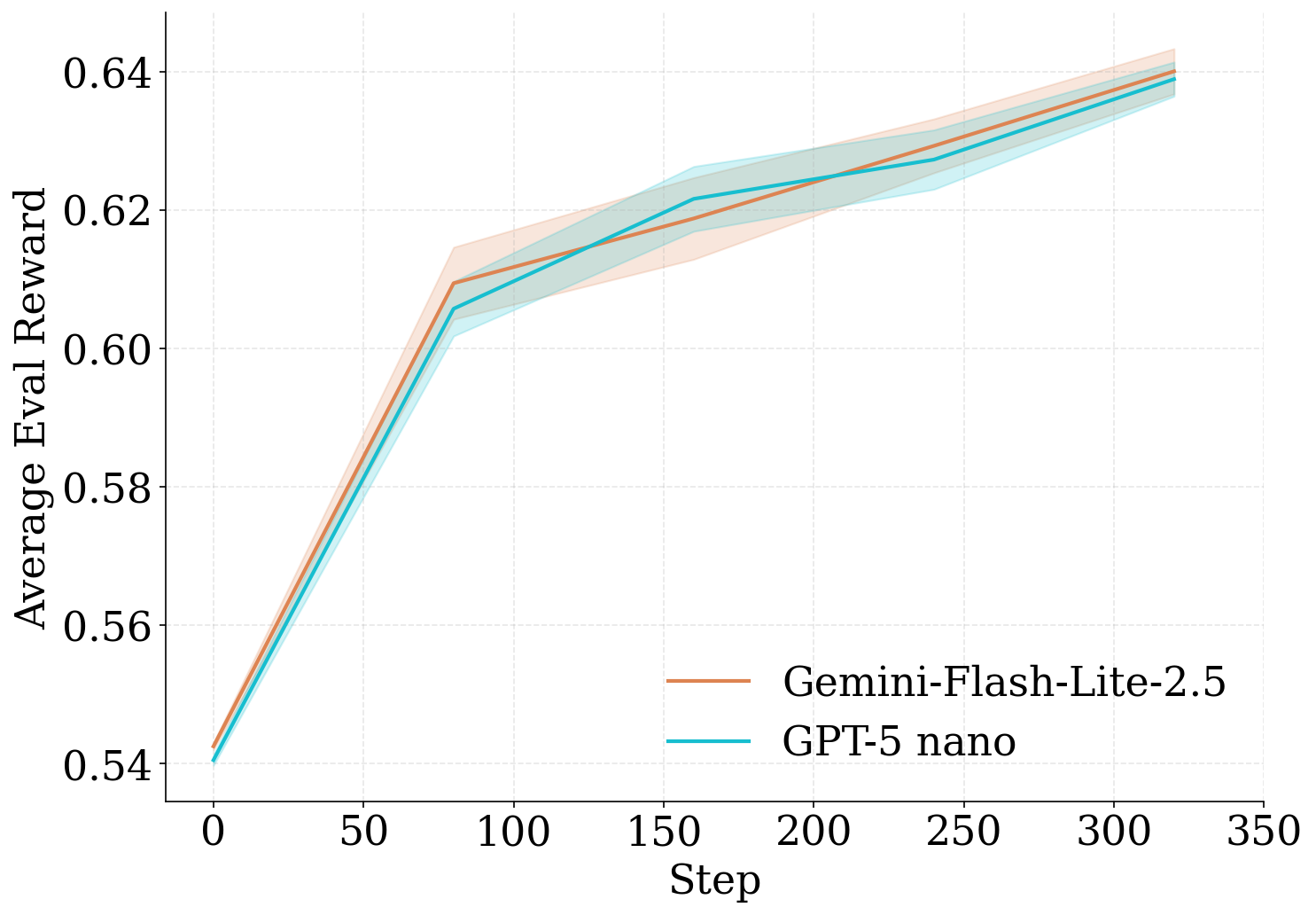}
        \caption{}
        \label{fig:mbe}
    \end{subfigure}
    \hfill
    \begin{subfigure}{0.325\textwidth}
        \centering
        \includegraphics[width=\textwidth]{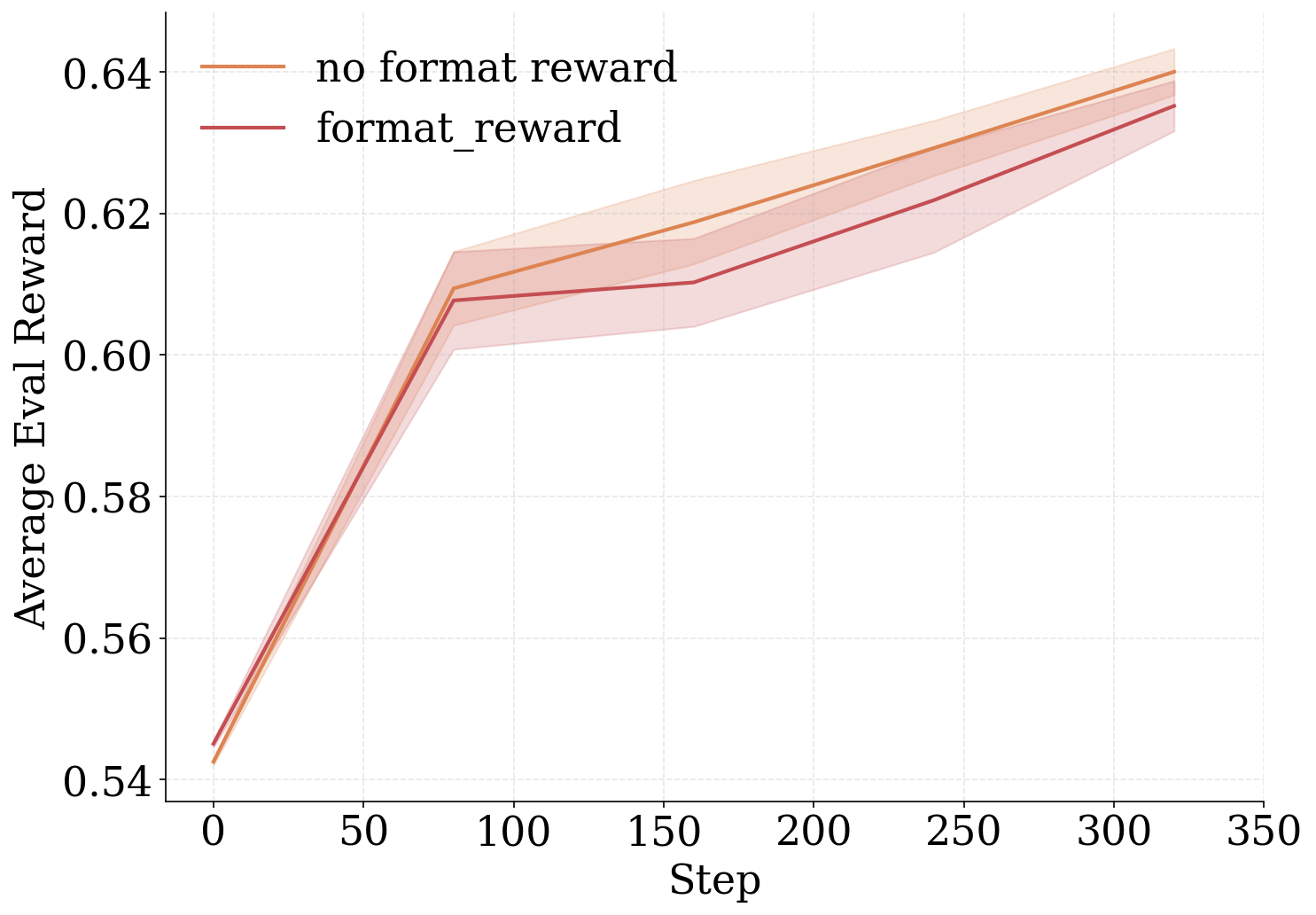}
        \caption{}
        \label{fig:format reward}
    \end{subfigure}
    \hfill
    \begin{subfigure}{0.325\textwidth}
        \centering
        \includegraphics[width=\textwidth]{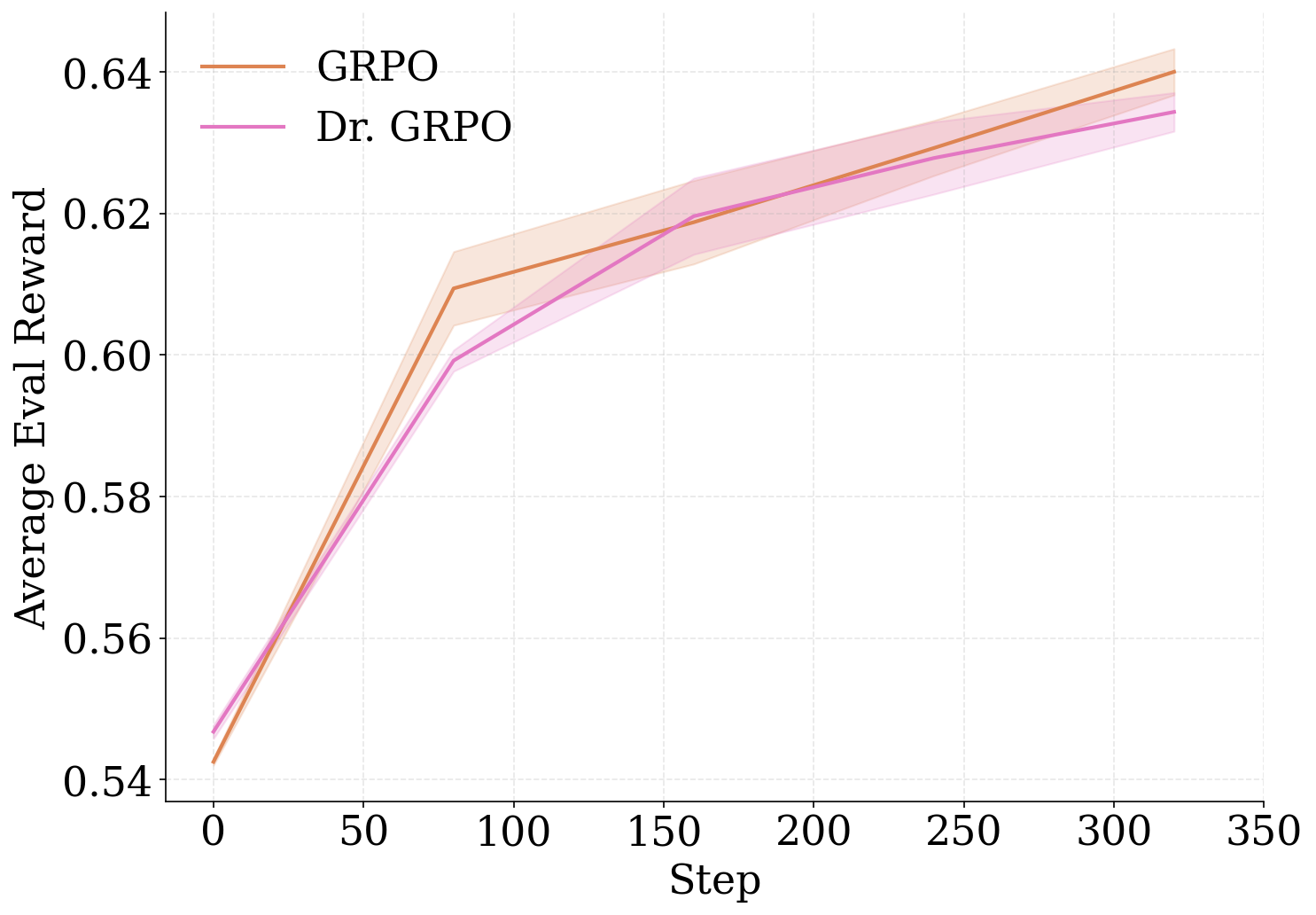}
        \caption{}
        \label{fig:dr grpo}
    \end{subfigure}
    \caption{a) The cheaper GPT-5 Nano LLM performs on par with Gemini-Flash-Lite-2.5 for assigning rewards based on predicted versus ground-truth answers.
b–c) Across the full training process, using format reward and Dr. GRPO shows no significant differences.}
    \label{fig:format reward and dr grpo}
\end{figure*}

In an additional experiment (see Figure~\ref{fig:dr grpo}), we compared GRPO against Dr. GRPO, and found that Dr. GRPO does not lead to clear performance gains over GRPO. 
That is, removing length normalization and advantage normalization does not lead to clear performance gains, implying that the on-policy nature of RLOO is the primary driver of the improvement. 
This observation is consistent with results in Figure 9 of \cite{liu2025understanding}. 
A deeper understanding of why keeping/removing these normalizations has a limited impact, particularly in interaction with the optimizer dynamics, remains an open question and is beyond the scope of this paper.

\begin{figure}[htbp]
    \centering
    \includegraphics[width=0.5\textwidth]{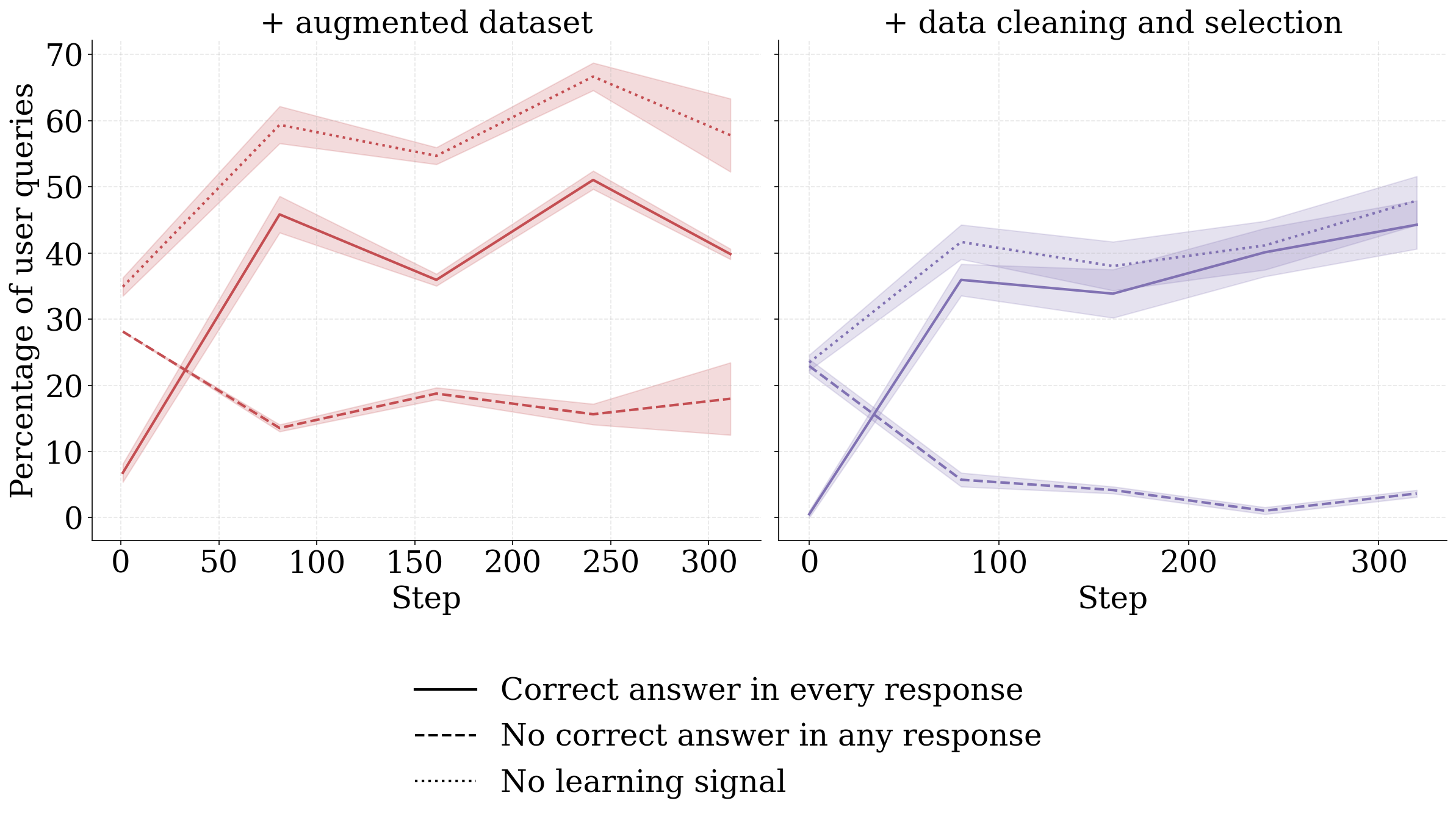}
    \caption{Data cleaning + selection significantly decreases user queries that do not provide learning signals. In GRPO or RLOO, if all episodes lead to the same rewards, all advantages are zero, so no learning signal is provided.}
    \label{fig:compare_prompt_rates}
\end{figure}

\subsection{Training Data Curation}
In addition, we show that curating the training dataset can further improve performance. 
We examine three complementary strategies: data augmentation, data cleaning, and data selection by difficulty.

\noindent\textbf{Data Augmentation.}
The previously used training dataset came from TQ, NQ, and 2Wiki. 
A close inspection of these benchmarks and GAIA, HLE, and BrowseComp show a significant hardness difference in the training and test queries.

To address this limitation, we augment the training data with an additional source, GenQAv4, introduced by \cite{2025mirorl}. 
This dataset contains substantially more challenging questions. 
We construct a new training dataset by mixing $12$k randomly sampled questions from the original dataset with $12$k questions from the GenQAv4 dataset. 

We performed an experiment illustrating the effect of data augmentation (red curves in Figure~\ref{fig:mbe_advantages}). 
The average evaluation reward curve shows that applying the data augmentation strategy slightly improves learning efficiency.  
However, the agent's behavior has changed significantly. 
The response length increases from around $4000$ tokens to $8000$ tokens. 
Further, the agent learns to generate many more search items with fewer search calls, indicating that it learns to search more items in a single web search tool call.

\noindent\textbf{Additional Data Cleaning.}
From the resulting data mixture, we filter out low-quality $(\text{question}, \text{reference answer})$ pairs using the same procedure applied when constructing the test dataset (see Appendix \ref{sec:cleaning_prompt}). 
This step removes a substantial fraction of noisy samples: $62\%$ of the original training split and $39\%$ of the GenQAv4 samples are excluded after cleaning. 

\noindent\textbf{Data Selection by Difficulty.}
Our final curation step selects questions based on their difficulty. 
This strategy is motivated by the observation that, for a considerable fraction of questions, all sampled episodes are either entirely correct or entirely incorrect (See Figure~\ref{fig:compare_prompt_rates}). 
Such questions provide no meaningful learning signal to RLOO because all advantages are zero. 

This phenomenon arises because some questions are trivial for the policy, while others are far beyond its current capability. 
To focus training on informative samples, we draw a large number of episodes (256) for each question in the mixture dataset and compute its \emph{answer correctness rate}, defined as the fraction of generated responses that yield a correct final answer. 
We retain only questions whose correctness rate falls within a predefined intermediate range.
Specifically, for GenQAv4, we keep questions with correctness rates between $2.5\%$ and $10\%$. 
For the original dataset, where questions are generally easier, and fewer samples fall into this narrow band, we use a wider range of $2.5\%$--$50\%$. 
After filtering, we retain $2{,}756$ questions from GenQAv4 and $2{,}756$ from the original dataset, resulting in a curated dataset of $5{,}512$ questions in total.

Experiments on the cleaned and curated dataset further improve the average evaluation reward (purple vs red). 
With this dataset, the agent performs web reads more frequently. 
Since web search returns only short snippets, whereas web reading accesses full webpage content (up to a truncation limit), increased web reading reflects deeper inspection of webpages rather than premature decisions based on search results.

\subsection{Error-Tolerant Test-Time Rollout}
Under the current formulation, an episode terminates immediately upon an error. 
Such errors include cases where the generated token sequence does not match the required pattern described in Section~\ref{sec:formulation}, the tool name is incorrect, or the tool-call script is malformed, etc. However, these errors do not necessarily require episode termination.

We conjecture that, if the policy is explicitly informed of the reason for a failure, it may be able to avoid repeating the same error in subsequent steps. 
Based on this hypothesis, we adopt an \emph{error-tolerant} rollout strategy at test time. 
Specifically, when an error is detected, the episode is not terminated. Instead, if an incorrectly formatted tool-call JSON script causes the error, we first attempt to automatically repair the script using a rule-based correction tool~\cite{Baccianella_JSON_Repair_-_2025}. If the repair attempt fails or if the error is of another type, we return an explicit error message to the agent, which then proceeds to take the next action, possibly fixing the error.

Under this error-tolerant rollout scheme, episodes continue until either the context-length limit or the maximum number of interaction turns is reached. 
We also experimented with applying this strategy during training, but did not observe meaningful improvements. 
At test time, we apply this strategy to an agent trained with all the above design choices. 
This error-tolerant rollout approach improves the average reward from $0.6986$ to $0.7107$. 
Average reward per data source can be found in \cref{tab:compare_with_other_systems}.

\section{Discussion and Conclusion}

This work provides a rigorous formulation of the deep research task as a reinforcement learning problem in an episodic MDP. 
Based on this formulation, we establish a competitive base agent. 
Built on this, we identify several key design choices that significantly improve the base agent's performance. 
With these choices, our final agent achieves state-of-the-art performance among deep research agents of comparable scale (7B) when evaluated across ten widely used benchmarks.

We also acknowledge several limitations of this work. 
First, our experiments rely on training and test data consisting of highly concrete questions with a small number of valid answers. 
In contrast, deep research agents should, in principle, accommodate much more general queries. 
In real-world settings, users often pose abstract questions with a vast space of possible answers of varying quality. 
Such questions pose substantially greater challenges for deep research agents, particularly for reward assignment, since it is infeasible to enumerate all valid answers and rewards must instead reflect nuanced answer quality. 
Second, the tools available to our agent are limited to web search and web reading, and do not support actions such as free-form browsing, interacting with on-page search interfaces, or logging into personal accounts. 
This limitation restricts the range of tasks the agent can effectively address. 
Third, our experiments are conducted exclusively with the \qwen model, leaving open the question of whether our findings generalize to other LLMs. 
Finally, all experiments are limited to a maximum of $320$ interaction steps. 
Although preliminary results indicate that training beyond this horizon can further improve performance, we do not systematically study longer training regimes due to the high cost of such experiments. 
Consequently, we do not evaluate asymptotic performance or whether relative performance trends persist with substantially more training. 
Our results should therefore be interpreted as reflecting performance under limited interaction budgets.

Despite these limitations, this work provides both a rigorous problem formulation and practical design insights for researchers and practitioners working on building deep research agents and, more broadly, LLM-based tool-using systems.

\bibliography{example_paper}
\bibliographystyle{icml2026}

\newpage
\appendix
\onecolumn
\noindent\textbf{\LARGE Appendix}

\section{Supplementary Details For Section \ref{sec:formulation}}\label{appendix:formulation}

\subsection{Description of GRPO}\label{sec:grpo}
GRPO~\citep{shao2024deepseekmath} can be seen as an extension of RLOO~\citep{kool2019reinforce} by introducing the following changes:
\begin{itemize}
    \item RLOO employs the current policy $\pi_\theta$ to generate data and performs the policy update immediately afterwards. Therefore, it is always on-policy. In contrast, GRPO uses $\pi_{\theta_\text{old}}$, which is updated every $l$ steps, to unroll trajectories --- the off-policy nature of GRPO results in an importance sampling ratio $\frac{\pi_\theta(A\mid S)}{\pi_{\theta_\text{old}}(A\mid S)}$ applied to the advantage terms.
    \item GRPO employs clipping to the importance sampling ratio to prevent aggressive updates to $\pi_\theta$ when it deviates too much from $\pi_{\theta_\text{old}}$, forming a trust region.
    \item GRPO normalizes the advantage by the returns' standard deviation, whereas RLOO does not.
    \item When optimizing $\theta$ on action $A$, the original form of GRPO does not leave the trajectory that depends on $A$ out of the group mean computation.
\end{itemize}
Therefore, we can borrow the derivation of RLOO in the MDP setting in Appendix~\ref{section: RLOO derivation} while keeping in mind the differences above.
In GRPO, at each step, a set of $m$ initial states $S_0^1, S_0^2, \dots, S_0^m$ is randomly sampled from the dataset. 
For each initial state, GRPO unrolls $n$ episodes following $\pi_{\theta_{\text{old}}}$, which is updated to the current policy $\pi_\theta$ every $l$ steps. 
Starting from $S_0^i$, the $n$ episodes can be denoted by $S_0^{i,j}, A_0^{i,j}, R_1^{i,j}, S_1^{i,j}, \dots, S_{T^{i,j}}^{i,j}, \forall j = 1, 2, \dots, n$.
GRPO then updates $\theta$ by the gradient of the following function:
\begin{align} \label{eq: GRPO}
L_{\text{GRPO}}(\theta)
\doteq &
 \frac{1}{m}
    \sum_{i=1}^m
    \frac{1}{n} \sum_{j=1}^n 
    \sum_{t=0}^{T^{i,j} - 1} 
\min \qty[ 
\frac{\pi_\theta \qty(A^{i,j}_{t} \mid S^{i,j}_t)}
{\pi_{\theta_{\text{old}}}\qty(A^{i,j}_{t} \mid S^{i,j}_{t})}
\hat{A}^{i,j}, \text{clip}
\qty(
\frac{\pi_\theta \qty(A^{i,j}_{t} \mid S^{i,j}_t)}
{\pi_{\theta_{\text{old}}}\qty(A^{i,j}_{t} \mid S^{i,j}_{t})}, 
1-\epsilon, 
1+\epsilon)
\hat{A}^{i,j}],
\end{align}
where $\hat{A}^{i,j} \doteq \frac{G^{i,j} - \bar{G}^i} {\delta + \sqrt{\frac{1}{n-1}\sum_{k=1}^n \qty(G^{i,k} - \bar{G}^i)^2}}, G^{i,j} \doteq \sum_{t=0}^{T^{i,j} - 1} R_{t+1}^{i,j},
\bar{G}^i \doteq \frac{1}{n}\sum_{k=1}^{n} G^{i,k}$, 
$\epsilon$ is the clipping hyperparameter, and $\delta$ is a small term to prevent division by zero. 
An additional KL-regularization term is used in the original GRPO but is not used in our work. 
In this work, we set $m = 64, n = 8$, and $ l = 8$. 

\subsection{Base Agent Fine-Tuning Experiment Details}\label{sec:base_agent_fine_tune}

The optimizer is AdamW \cite{loshchilov2017decoupled} with a learning rate of $1e-6$ and $\beta_1 = 0.9, \beta_2 = 0.999$ and the regularization coefficient being $0.01$. We used a gradient norm clipping with a threshold of $0.2$. We used mixed-precision training, where parameters and optimizer states are stored in FP32 and temporary values such as gradients and activations are stored in FP16. We truncate the response from Jina reader to $20,000$ tokens before sending to the LLM summarization model. Web read responses for different webpages are processed separately by the LLM summarization model.

\subsection{Detailed Descriptions of Benchmarks}\label{sec:benchmarks}
\begin{table}[!htbp]
     \small
    \centering
    \begin{tabular}{p{0.2\textwidth}p{0.5\textwidth}p{0.2\textwidth}}
    \toprule
         \textbf{Benchmark} & \textbf{Question} & \textbf{Ground-Truth}\\ 
         \midrule
         NQ & When did the first tourist travel into space? & April 28 , 2001 \\
          \midrule
         TQ & British monarch Henry VIII was born in which royal palace? &  Greenwich Palace Palace of Placentia\\
          \midrule
         PopQA & Who is the author of Gemma Doyle Trilogy? & Libba Bray \\
          \midrule
         HotpotQA & What is the name of this public research university in the U.S. state of Illinois, where Robert McKim is Professor of Religion and Professor of Philosohpy? & University of Illinois at Urbana-Champaign \\
          \midrule
         MuSiQue & What is the capital of the county which contains Hickory Grove Estates, Mississippi? & Starkville\\
         \midrule
         2Wiki & Who is Princess Anna Elisabeth Louise Of Brandenburg-Schwedt's maternal grandfather? & Frederick William I of Prussia\\
         \midrule
         Bamboogle & What rocket was the first spacecraft that ever approached Uranus launched on? & Titan IIIE\\
          \midrule
         HLE & When viewed as matrices, which flags of African nations have the same linear algebraic rank as the flag of Denmark? Assume that values in each matrix are chosen so that the rank of each flag is maximal. & Benin and Madagascar\\
          \midrule
         GAIA & How many at bats did the Yankee with the most walks in the 1977 regular season have that same season? & 519\\
         \midrule
         BC & In what year did the event occur that led to the loss of lives and the dedication of a monument in their honor which was constructed prior to 1970 in former Yugoslavia in one of the top 4 largest cities in Bosnia per the 2013 population census and by an artist who was born in 1928? & 1942\\
         \bottomrule
    \end{tabular}
    \caption{Sample instances from the evaluation benchmarks.}
    \label{tab:benchmark examples}
\end{table}
In this section, we provide detailed descriptions of the queries included in each benchmark used.
\begin{itemize}
    \item \textit{Natural Questions (NQ)} \cite{kwiatkowski2019natural}: A large-scale question-answering dataset derived from real Google search queries, testing the agent's ability to answer factoid questions using Wikipedia articles.
    
    \item \textit{TriviaQA} \cite{joshi2017triviaqa}: A reading comprehension dataset containing trivia questions paired with evidence documents, evaluating the agent's capacity to locate and extract relevant information from web sources.
    
    \item \textit{PopQA}: A dataset focused on questions about popular entities and topics, assessing the agent's performance on queries requiring up-to-date knowledge from current web content.
    
    \item \textit{HotpotQA} \cite{yang2018hotpotqa}: A multi-hop reasoning dataset requiring the agent to gather and synthesize information from multiple documents to answer complex questions.
    
    \item \textit{2WikiMultiHopQA} \cite{ho2020constructing}: A challenging benchmark designed specifically for multi-hop reasoning over Wikipedia, where answering requires connecting information across multiple articles.
    
    \item \textit{Musique} \cite{trivedi2022musique}: A multi-hop question-answering benchmark that tests compositional reasoning abilities, requiring the agent to perform sequential information gathering and inference steps.
    
    \item \textit{Bamboogle (BAMB)} \cite{press2022measuring}: A dataset containing questions that cannot be answered using the model's parametric knowledge alone, necessitating active web search and information retrieval.
    
    \item \textit{GAIA} \cite{mialon2023gaia}: A benchmark presenting real-world complexity with sophisticated reasoning chains, evaluating the agent's ability to handle realistic, challenging research tasks.

    \item \textit{BrowseComp} \cite{wei2025browsecomp}: A standardized evaluation suite for web browsing competency, testing the agent's ability to navigate and extract information from dynamic web pages across multiple languages.

    \item \textit{Human's Last Exam} \cite{phan2025humanity} A comprehensive benchmark assessing an agent’s general reasoning, factual recall, and multi-domain understanding, serving as a holistic test of advanced language and reasoning capabilities.
\end{itemize}
Sample instances from these benchmarks are provided in Table \ref{tab:benchmark examples}.

\subsection{Dataset Cleaning}\label{sec:cleaning_prompt}
In our evaluation benchmarks, some of the questions and the corresponding answers could be of low quality. Common issues include questions that are not well-formed (\eg, fragmented statements or incomplete sentences), ambiguous questions for which the provided reference answers do not cover all valid answers, and cases where some of the reference answers themselves are incorrect. In Table \ref{tab:low_quality_example}, we provide some examples.
\begin{table}[htbp]
  \centering
  \small
  \begin{tabular}{p{0.25\textwidth}p{0.2\textwidth}p{0.45\textwidth}}
  \toprule
  \textbf{Question} & \textbf{Answer} & \textbf{Reason for Low Quality} \\
  \midrule
  Where in the country where Bodindecha was born was The Beach filmed? & island Koh Phi Phi & The provided answer is not the only correct answer. While Koh Phi Phi was a primary and iconic location, scenes for \textbf{The Beach} were also filmed in other parts of Thailand, including Phuket, Bangkok, and Khao Yai National Park. Therefore, ``island Koh Phi Phi'' is an incomplete answer as other valid answers exist. \\
  \midrule
  What other film is the cast member of Now You See Him, Now You Don't a character for? & The Hateful Eight & The question, "What other film is the cast member of Now You See Him, Now You Don't a character for?" is a complete sentence. However, it is fundamentally ambiguous because it doesn't specify \textbf{which} cast member of the film it is referring to. A film has multiple cast members, and each has their own filmography. \\
  \bottomrule
  \end{tabular}
  \caption{Examples of low-quality question-answer pairs identified during data cleaning.}
  \label{tab:low_quality_example}
\end{table}

\subsection{Configuration Differences Among Different Agents}\label{sec:config}

All the agents considered in this work are initialized from Qwen2.5-7B-Base, with the exception of DeepResearcher, which utilizes Qwen2.5-7B-Instruct. This choice is necessitated by model availability and performance: R1-Searcher and ASearcher are only released as Base-trained versions, whereas DeepResearcher is exclusively available as an Instruct-trained model. For agents where both versions are available, Search-R1 and ZeroSearch, the respective authors report that the Qwen2.5-7B-Base initialization outperforms the Instruct variant. In their own experimental study presented in the paper, R1-Searcher is evaluated under an offline search setting. with the exception of the Bamboogle benchmark, which is tested on a live web search environment, Search-R1 is also assessed exclusively in an offline search setting. ZeroSearch and DeepResearcher are evaluated using live web search, whereas ASearcher is evaluated under both offline and online (live web) search conditions.

Another distinction among these agents lies in whether they use live web access or static/simulated web data during training. Similarly to our base agent, ASearcher also utilizes Serper for web search and Jina for page retrieval. DeepResearcher employs Serper alongside a text-based browser (comparable to Lynx~\cite{lynx}) for content retrieval. R1-Searcher utilizes a static KILT~\citep{petroni2021kilt} Wikipedia index rather than live web access. Search-R1 interleaves reasoning steps with retrieval actions but restricts its search space to a local Wikipedia retriever. ZeroSearch bypasses live web calls entirely during RL training, instead utilizing an LLM to simulate search engine responses.

\section{Supplementary Details For Section 3}
\label{appendix: section 3}

\subsection{Rule-Based Rewards}\label{sec:metric}
For completeness, we include the definition of the F1 score and Exact Match here.
\begin{itemize}
        \item \textit{F1 Score:} The F1 score measures the harmonic mean of precision and recall between the set of tokens in the generated answer and the ground-truth. Before comparison, both texts are normalized by converting to lowercase and removing punctuation. Let $G$ and $T$ denote the sets of tokens of the generated answer and the ground-truth, respectively. Define $C \doteq G \cap T$. Then, precision is defined as $P \doteq \abs{C}/\abs{G}$, and recall is defined as $R \doteq \abs{C}/\abs{T}$. We compute the word-level F1 score as follows,
        \begin{align}
        F_1(G, T) \doteq \frac{2PR}{P + R}.
        \end{align}
        This approach provides a nuanced assessment of content overlap, rewarding answers that are substantially correct even if they are not lexically identical to the ground-truth. 
        \item \textit{Exact Match:} The Exact Match (EM) reward is a stricter evaluation metric. This binary measure awards a score of 1 if the normalized predicted answer is identical to any of the ground-truth answers, and 0 otherwise. Although less flexible than the F1 score, it serves as a clear indicator of complete accuracy.
    \end{itemize}
    
\subsection{Limitations of Token-Level F1 Evaluation}
\label{sec:f1-limitations}

We compare F1 scores against LLM-based evaluation (Gemini Flash) across 1176 question-answer pairs and identify two categories of failures: false positives (incorrect answers rewarded) and false negatives (correct answers penalized).

\subsubsection{False Positives: F1 Rewards Incorrect Answers}

\paragraph{Pattern 1: Mathematical Notation Confusion.}
F1's token-based matching cannot distinguish between mathematically distinct objects that share symbolic components. See Table 4.

\begin{table}[h]
\small
\centering
\begin{tabular}{p{4.5cm}p{3.5cm}p{3.5cm}c}
\toprule
\textbf{Question (abbreviated)} & \textbf{Ground Truth} & \textbf{Predicted} & \textbf{F1} \\
\midrule
Topological invariant group for 2D free fermion model? & \texttt{\$2\textbackslash mathbb\{Z\}\$} & \texttt{\$\textbackslash mathbb\{Z\}\_2\$} & 1.00 \\
Compute $\inf_{f\in S}f(\pi)$. & \texttt{\textbackslash frac\{1-1/(\textbackslash pi+1)\}} \texttt{\{\textbackslash log(\textbackslash pi+1)\}} & \texttt{\textbackslash frac\{1\}\{\textbackslash pi+1\}} & 0.86 \\
Poincar\'{e} polynomial of $\mathfrak{g}$? & \texttt{1 + 3x + 6x\^{}2 + 8x\^{}3 + ...} & \texttt{1 + x + x\^{}2 + x\^{}3 + ...} & 0.82 \\
\bottomrule
\end{tabular}
\caption{F1 incorrectly rewards mathematically distinct expressions. $2\mathbb{Z}$ (even integers) and $\mathbb{Z}_2$ (integers mod 2) are fundamentally different algebraic structures.}
\end{table}

\paragraph{Pattern 2: Factually Incorrect Values.}
Token overlap between correct and incorrect answers produces misleadingly high F1 scores. See Table 5. 

\begin{table}[h]
\small
\centering
\begin{tabular}{p{3.5cm}p{3.5cm}p{3.5cm}c}
\toprule
\textbf{Question (abbreviated)} & \textbf{Ground Truth} & \textbf{Predicted} & \textbf{F1} \\
\midrule
Composer's date of birth? & \texttt{2 June 1943} & \texttt{June 3, 1943} & 0.67 \\
Director's birthday? & \texttt{June 16, 1911} & \texttt{June 16, 1909} & 0.67 \\
Oz book illustrator? & \texttt{Lauren McGraw Wagner} & \texttt{Laurel McGraw Wagner} & 0.67 \\
\bottomrule
\end{tabular}
\caption{F1 fails to detect wrong dates (2 vs 3, 1911 vs 1909) and wrong names (Lauren vs Laurel).}
\end{table}

\subsubsection{False Negatives: F1 Penalizes Correct Answers}

\paragraph{Pattern 1: Number Formatting and Representation.} See Table 6.

\begin{table}[h]
\small
\centering
\begin{tabular}{p{4cm}p{2.5cm}p{2.5cm}c}
\toprule
\textbf{Question (abbreviated)} & \textbf{Ground Truth} & \textbf{Predicted} & \textbf{F1} \\
\midrule
2020 population of island? & \texttt{56000} & \texttt{56,000} & 0.00 \\
How many Wikipedia edits? & \texttt{2732} & \texttt{2,732} & 0.00 \\
River length? & \texttt{1472 km} & \texttt{1,472 kilometres} & 0.00 \\
Which stanza? & \texttt{2} & \texttt{Second} & 0.00 \\
\bottomrule
\end{tabular}
\caption{F1 fails on number formatting (comma separators) and representation (digits vs words, cardinal vs ordinal).}
\end{table}

\clearpage
\paragraph{Pattern 2: Unicode and Format Variations. } See Table 7. 

\begin{table}[h]
\small
\centering
\begin{tabular}{p{4cm}p{2.5cm}p{2.5cm}c}
\toprule
\textbf{Question (abbreviated)} & \textbf{Ground Truth} & \textbf{Predicted} & \textbf{F1} \\
\midrule
Circuit complexity upper bound? & \texttt{TC\textsuperscript{0}} & \texttt{TC0} & 0.00 \\
How many DFA states? & \texttt{D} & \texttt{4} & 0.00 \\
BAFTA Best Actor 2015? & \texttt{eddieredmayne} & \texttt{Eddie Redmayne} & 0.00 \\
Wolf pack leader name? & \texttt{akele} & \texttt{Akela} & 0.00 \\
\bottomrule
\end{tabular}
\caption{F1 fails on Unicode superscripts, multiple choice formats, and spelling/spacing in names.}
\end{table}
\paragraph{Pattern 3: Semantic Equivalence.} See Table 8.

\begin{table}[h]
\small
\centering
\begin{tabular}{p{3.5cm}p{3.5cm}p{3.5cm}c}
\toprule
\textbf{Question (abbreviated)} & \textbf{Ground Truth} & \textbf{Predicted} & \textbf{F1} \\
\midrule
Why no Soviet medals 1984? & \texttt{their nations boycotted games} & \texttt{Boycott} & 0.00 \\
Who makes decisions in autocracy? & \texttt{one person} & \texttt{Autocrat} & 0.00 \\
Who did Ares side with? & \texttt{the Trojan side} & \texttt{Trojans} & 0.00 \\
How did Dunkin' Donuts help? & \texttt{a commercial} & \texttt{Through advertisements} & 0.00  \\
\bottomrule
\end{tabular}
\caption{F1 fails to recognize semantic equivalence: verbose vs concise, generic vs specific terms, synonymous expressions.}
\end{table}

\subsection{Validating AI Feedback as Training Reward}
\label{sec:ai-feedback-validation}

Having established the superiority of LLM-based evaluation over token-level F1, we now validate that cost-effective LLMs can reliably generate training rewards. We compare Gemini Flash Lite (our reward model) against Gemini Flash (serving as a proxy for human judgment). Across 1,176 evaluation instances, the two models achieved a 97.62\% agreement rate (1,148 agreements, 28 disagreements).

We manually analyzed all 28 disagreement cases and categorized them into three types: (1) Flash Lite incorrectly lenient (4 cases), (2) Flash incorrectly strict (13 cases), and (3) Flash Lite incorrectly strict (11 cases). Overall, among the 28 cases where the two LLMs disagree, Flash Lite is incorrect in 4 + 11 = 15 cases, while Flash is incorrect in 13 cases. These results indicate that, despite being substantially cheaper than Flash, Flash Lite exhibits a comparable capability in judging the correctness of a predicted answer given a ground-truth answer. This suggests that very inexpensive LLMs may be sufficient for answer-judging tasks.

\subsection{REINFORCE Leave-One-Out Derivation}
\label{section: RLOO derivation}
Let $\mathcal{X}$ and $\mathcal{Y}$ be general finite sample spaces and $f$ be a mapping from $\mathcal{X} \times \mathcal{Y}$ to a scalar.
Define $\rho: \mathcal{X} \to [0,1]$ as a probability mass function over $\mathcal{X}$.
Let $p_\theta: \mathcal{Y} \times \mathcal{X} \to [0,1]$ be a conditional probability mass function parameterized by $\theta$.
Given that we wish to estimate the gradients $\nabla_\theta \E_{X \sim \rho, Y \sim p_\theta(\cdot \mid X)}\qty[f(X,Y)]$, the REINFORCE with baseline estimator~\citep{williams1992simple} allows us to relate the gradients with the expectation
\begin{align}
\label{eq: policy gradient}
    \nabla_\theta \E_{X \sim \rho, Y \sim p_\theta(\cdot \mid X)}\qty[f(X,Y)]
    = \E_{X \sim \rho}\qty[\E_{Y \sim p_\theta(\cdot \mid X)} \qty[\nabla_\theta \log p_\theta(Y \mid X) \qty(f(X, Y) - B)]],
\end{align}
where $B$ is a scalar baseline independent of $Y$.
Suppose we have i.i.d. samples $X_1, X_2, \dots, X_m$ sampled from $\rho$, and, for each $X_i$, $Y^i_1, Y^i_2, \dots, Y^i_n$ sampled i.i.d. from $p_\theta(\cdot \mid X_i)$, 
then we can construct an unbiased estimator of the gradient in ~\eqref{eq: policy gradient} as
\begin{align}
\label{eq: REINFORCE estimator with baseline}
    \frac{1}{m}\sum_{i = 1}^m \frac{1}{n}\sum_{j=1}^n \nabla_\theta \log p_\theta\qty(Y^i_j \mid X_i)\qty(f\qty(X_i, Y^i_j) - B^i_j),
\end{align}
where $B^i_j$ is independent of $Y^i_j$.
\citet{kool2019reinforce} show that we can construct $B^i_j$ as $B^i_j = \frac{1}{n-1} \sum_{k \neq j} f(X_i, Y^i_k)$
and~\eqref{eq: REINFORCE estimator with baseline} becomes
\begin{align}
    &\frac{1}{m}\sum_{i=1}^m \frac{1}{n} \sum_{j=1}^n
    \nabla_\theta \log p_\theta\qty(Y^i_j \mid X_i)\qty(f\qty(X_i, Y^i_j) - \frac{1}{n-1} \sum_{k\neq j} f\qty(X_i, Y^i_k))\\
    \label{eq: rloo estimator} 
    =& \frac{1}{m}\sum_{i=1}^m\frac{1}{n-1} 
    \sum_{j=1}^n \nabla_\theta \log p_\theta\qty(Y^i_j \mid X_i) 
    \qty(f\qty(X_i, Y^i_j) - \frac{1}{n}\sum_{k=1}^n f\qty(X_i, Y^i_k)).
\end{align}
Estimator~\eqref{eq: rloo estimator} is known as the REINFORCE Leave-One-Out (RLOO) estimator, and~\citet{kool2019reinforce} proved its unbiasedness without conditioning on $X$.
Our setting is a straightforward extension of their result.

We now derive the RLOO objective in the MDP scenario using this estimator.
Define $\tau \doteq S_0, A_0, R_1, S_1, \dots, S_T$ as an unrolled trajectory under the parameterized policy $\pi_\theta$.
We map $X$ and $Y$ to different portions of $\tau$.
Specifically, we let $X \doteq S_0$ be the initial state and let $Y \doteq A_0, R_1, S_1, \dots, S_T$ be the rest of the trajectory following $S_0$.
Then, we have $\rho = \mu$ and 
\begin{align}
    p_\theta(Y \mid X) 
    =&
    \prod_{t=0}^{T-1}\pi_\theta\qty(A_t \mid S_t)p\qty(S_{t+1} \mid S_t, A_t)\\
    \log p_\theta(Y \mid X) 
    =& \sum_{t = 0}^{T-1} 
    \qty(\log \pi_{\theta}(A_t \mid S_t) + \log p(S_{t+1} \mid S_t, A_t)).
\end{align}
Hence, it holds that
\begin{align}
\label{eq: p_theta in pi}
    \nabla_\theta \log p_\theta(Y \mid X) = \sum_{t=0}^{T-1} \nabla_\theta \log \pi_\theta(A_t \mid S_t).
\end{align}
We further define $f(X,Y) \doteq \sum_{t=0}^{T-1} R_{t+1}$.
Intuitively, $f$ here simply maps an episode to its return as the sum of the immediate rewards.
Suppose we have sampled $m$ i.i.d. initial states $X_1, X_2, \dots, X_m$ from $\rho$, where $X_i = S_0^i$, and, for each $X_i$, $n$ i.i.d. episodes $Y^i_1, Y^i_2, \dots, Y^i_n$ from $p_\theta(\cdot \mid X_i)$, 
where $Y^i_j = A_0^{i,j}, R_1^{i,j}, S_1^{i,j}, \dots, S_{T^{i,j}}^{i,j}$,
we have by~\eqref{eq: rloo estimator} and~\eqref{eq: p_theta in pi} that
\begin{align}
    &\nabla_\theta \E_{X\sim \rho, Y \sim p_\theta(\cdot \mid X)}\qty[f(X,Y)]\\
    =&
    \E_{\rho, p_\theta}\qty[
    \frac{1}{m}\sum_{i=1}^m
    \frac{1}{n-1} \sum_{j=1}^n 
    \sum_{t=0}^{T^{i,j} - 1} \nabla_\theta \log \pi_\theta\qty(A^{i,j}_t \mid S^{i,j}_t) 
    \qty(f\qty(X_i, Y^i_j) - \frac{1}{n}\sum_{k=1}^n f\qty(X_i, Y^i_k))
    ].
\end{align}
Thus, rewriting the notation, we have
\begin{align}
     \nabla_\theta \E_{\pi_\theta}\qty[\sum_{t=0}^{T-1} R_{t+1}]
     =
     \E_{\pi_\theta}\qty[
     \frac{1}{m}
     \sum_{i=1}^m
     \frac{1}{n-1} \sum_{j=1}^n 
     \sum_{t=0}^{T^{i,j} - 1} \hat{A}^{i,j} \nabla_\theta \log \pi_\theta\qty(A^{i,j}_t \mid S^{i,j}_t)
     ],
\end{align}
where $\hat{A}^{i,j} \doteq G^{i,j} - \bar{G}^i$, $\bar{G}^i \doteq \frac{1}{n} \sum_{k=1}^n G^{i,k}$, and
$G^{i,j} \doteq \sum_{t=0}^{T^{i,j} - 1} R_{t+1}^{i,j}$.
We hence define the RLOO objective as
\begin{align}
    L_{\text{RLOO}}(\theta)
    \doteq 
    \frac{1}{m}
    \sum_{i=1}^m
    \frac{1}{n-1} \sum_{j=1}^n 
    \sum_{t=0}^{T^{i,j} - 1} \hat{A}^{i,j} \log \pi_\theta\qty(A^{i,j}_t \mid S^{i,j}_t).
\end{align}

\begin{table}[ht]

\centering
\begin{tabular}{lccccc}
\hline
\textbf{Method} & \textbf{2WIKI} & \textbf{TQ} & \textbf{NQ} & \textbf{BAM} & \textbf{POP} \\
\hline
R1-Searcher & 61.6$\pm$1.3 & 65.0$\pm$0.7 & 66.2$\pm$0.4 & 62.4$\pm$1.7 & 65.1$\pm$0.9 \\
Search-R1 & 78.4$\pm$1.1 & 74.2$\pm$0.5 & 79.2$\pm$1.2 & 75.3$\pm$2.2 & 77.2$\pm$0.4 \\
ZeroSearch & 17.6$\pm$0.3 & 31.4$\pm$0.5 & 30.0$\pm$0.4 & 53.9$\pm$1.4 & 39.7$\pm$1.3 \\
ASearcher & 84.4$\pm$1.0 & 84.6$\pm$1.1 & 87.2$\pm$0.3 & 74.4$\pm$1.7 & 81.9$\pm$0.2 \\
DeepResearcher & 85.4$\pm$0.5 & 79.8$\pm$0.3 & 89.6$\pm$0.9 & 78.3$\pm$1.5 & 81.1$\pm$0.5 \\ 
WebSailor & 88.8 $\pm$ 0.8  & \textbf{92.8} $\pm$ 0.4 & 97.6 $\pm$ 0.4 & 86.8 $\pm$ 0.2 & \textbf{87.9} $\pm$ 0.2 \\
\hline

Base agent & 92.0 $\pm$ 1.2 & 82.7 $\pm$ 1.4 & 88.2 $\pm$ 0.3 & 84.3 $\pm$ 0.7 & 83.6 $\pm$ 0.7 \\
Our Best Agent & \textbf{90.8} $\pm$ \textbf{0.2}  & 92.6 $\pm$ 0.5& \textbf{97.8} $\pm$ \textbf{0.7} & \textbf{92.8} $\pm$ \textbf{1.0} & 86.3 $\pm$ 0.9\\

\hline
\end{tabular}

\vspace{2mm}
\centering
\begin{tabular}{lcccccc}
\hline
\textbf{Method} & \textbf{MUS} & \textbf{HOT} & \textbf{HLE} & \textbf{GAIA} & \textbf{BC} & \textbf{AVG} \\
\hline
R1-Searcher & 51.5$\pm$2.2 & 62.6$\pm$0.2 & 4.1$\pm$0.0 & 4.9$\pm$0.7 & 0.8$\pm$0.0 & 40.8 $\pm$ 0.3\\
Search-R1 & 61.0$\pm$0.4 & 72.8$\pm$0.3 & 11.1$\pm$0.6 & 18.7$\pm$1.1 & 0.6$\pm$0.4 & 50.9 $\pm$ 0.4\\
ZeroSearch & 11.4$\pm$0.9 & 13.8$\pm$0.2 & 7.0$\pm$0.4 & 8.4$\pm$0.5 & 0.4$\pm$0.4 & 18.8 $\pm$ 0.2\\
ASearcher & 64.9$\pm$1.5 & 84.8$\pm$1.8 & 11.4$\pm$1.1 & 16.9$\pm$0.8 & 2.6$\pm$0.2 & 57.6 $\pm$ 0.3\\

DeepResearcher & 62.8$\pm$0.6 & 79.8$\pm$0.7 & 10.2$\pm$0.4 & 20.6$\pm$0.9 & 2.2$\pm$0.4 & 56.6 $\pm$ 0.3\\ 
WebSailor & 69.0 $\pm$ 3.5 & \textbf{92.8} $\pm$ 0.3 & 12.8 $\pm$ 3.4 & 34.0 $\pm$ 0.5 & 5.6 $\pm$ 1.2 & 66.8 $\pm$ 0.6 \\
\hline
Base agent  & 67.0 $\pm$ 2.6 & 80.3 $\pm$ 1.9 & 9.6 $\pm$ 0.5 & 26.2 $\pm$ 2.8 & 1.9 $\pm$ 1.9 & 61.4 $\pm$ 0.5 \\
Our best agent & \textbf{81.0} $\pm$ \textbf{2.3} & 92.0 $\pm$ 0.3 & \textbf{17.6} $\pm$ \textbf{0.3} & \textbf{49.2} $\pm$ \textbf{2.5} & \textbf{6.2} $\pm$ \textbf{0.9} & \textbf{71.07} $\pm$ \textbf{0.4}\\

\hline
\end{tabular}
\caption{Performance comparison of considered deep research agents across prevalent QA benchmarks. We report the means and standard errors across 4 independent runs.}
\label{tab:qa_comparison_updated}
\end{table}

\newpage
\section{Prompts} \label{appendix:prompts}
\begin{promptbox}[System Prompt]
Today is <DATE TO BE FILLED IN>.

You are a deep research assistant capable of performing iterative, evidence-based research to answer complex factual questions.

You must always produce one of the two types of outputs.

### Output Type 1 - when making a tool call:

Output Type 1 Format:

<think>
INSTRUCTIONS FOR WRITING THE THINK CONTENT WILL BE GIVEN SHORTLY.
</think>
<tool_call>
INSTRUCTIONS FOR WRITING THE TOOL_CALL CONTENT WILL BE GIVEN SHORTLY.
</tool_call>

### Output Type 2 - when giving the answer:

Output Type 2 Format:

<think>
INSTRUCTIONS FOR WRITING THE THINK CONTENT WILL BE GIVEN SHORTLY.
</think>
<answer>
INSTRUCTIONS FOR WRITING THE ANSWER CONTENT WILL BE GIVEN SHORTLY.
</answer>

Instructions for writing THINK CONTENT:
    step 1:
        - If no tool has been called:
            analyze the question and determine all knowns and output your overall research plan.
        - If a web_read tool response is received:
            summarize the knowns that you have learned from the response.
        - If a web_search tool response is received:
            The web_search responses are snippets of the actual content of webpages. They are NOT reliable and can not be used to answer the question. You should use the web_read tool to fetch the actual content of the webpages.
    
    step 2:
        - Regardless of whether a tool response is received, try to derive new knowns from all existing knowns using logical deduction or mathematical calculations.

    step 3:
        - Based on all existing knowns, identify any remaining gaps in knowledge. If so, explain how you plan to fill the gaps. Otherwise, explain how you derived the answer from the knowns.

Instructions for writing TOOL_CALL CONTENT:
    - The tool_call block should be a tool call dictionary of the following format: {{"name": TOOL_NAME, "arguments": TOOL_ARGUMENTS}}
    - Additional instruction on the details of the available tools will be given shortly.

Instructions for writing ANSWER CONTENT:
    - Only include your final answer to the question. No explanations, reasoning, citations, etc.
\end{promptbox}

\begin{promptbox}[Prompt for postprocess Jina web read results]
You are a helpful assistant. I will provide you:
    * A question and a webpage.

    You should generate a response as follows:
    1) Read the webpage carefully to answer the question. If the webpage content does not contain information needed to answer the question, explain why. 
    2) If you provide an answer, please also provide detailed information, such as citing certain content from the webpage content, that backs up the answer.
    
    Return the response.
\end{promptbox}

\begin{promptbox}[Evaluation Prompt]
You are an expert LLM judge. Given a user QUESTION, one or more GROUND_TRUTHS, and a PREDICTION, decide if the PREDICTION is correct.

Inputs:
- QUESTION: free text. Enclosed in <question>...</question> tags.
- GROUND_TRUTHS: one or more reference answers, separated by the literal token <|answer_split|>. Enclosed in <ground_truth>...</ground_truth> tags.
- PREDICTION: the model's answer. Enclosed in <agent_prediction>...</agent_prediction> tags.

General rule:
- Answer is CORRECT only if the PREDICTION is semantically equivalent to at least one GROUND_TRUTH for what the QUESTION asks. Otherwise, answer is INCORRECT. Do not award partial credit.
- give your rationale based on the detailed Evaluation principles below.

Evaluation principles (apply in order):
1) Scope: Judge only what the QUESTION requires based on GROUND_TRUTHS. Ignore extraneous but non-contradictory text; penalize explicit contradictions.
2) An empty answer is INCORRECT automatically.
3) Semantic equivalence: Allow paraphrases, synonyms, reordering, casing, punctuation, and minor typos.
4) Numbers \& units: Treat numerically equivalent values as equal, including:
   - Exact integers/floats; computed values (e.g., 1/2 == 0.5).
   - Unit conversions (e.g., 1 km == 1000 m).
   - Tolerance: if the QUESTION does not demand exactness, accept within $\pm1\%$ OR an absolute tolerance of $\pm$1e-6 for small numbers. If the QUESTION says ``exact/rounded to N", require that.
5) Dates/times: Accept format variants (YYYY-MM-DD, "Jan 5, 2024"), time-zone-free matches, and week/day names if unambiguous.
6) Lists/sets:
   - If the QUESTION asks for a single item, matching any one correct GROUND_TRUTH is sufficient.
   - If it asks for multiple items/all items, require the same set content in GROUND_TRUTHS (order-agnostic) unless the QUESTION specifies order.
7) Multi-span answers: If the GROUND_TRUTH is multi-part, the PREDICTION must correctly provide all required parts.
8) Hedging/uncertainty: Answers that are uncertain ("maybe", "I think", "not sure") or explicitly partial count as INCORRECT.
9) Refusals/safety: If the correct behavior is to refuse (e.g., the GROUND_TRUTH is "refuse" or "cannot determine"), mark refusal as CORRECT when the PREDICTION appropriately refuses for the right reason.
10) Hallucinations: If the PREDICTION adds concrete claims that conflict with the GROUND_TRUTHS, mark INCORRECT.
11) Code/math form: Algebraically equivalent expressions are acceptable (e.g., (x+1)^2 vs x^2+2x+1). For booleans, accept true/false/yes/no equivalents.
12) Name: For named entities (people, places, organizations), accept common abbreviations, acronyms, and alternate spellings.

Notes:
- Do NOT reveal or quote ground truths verbatim in the rationale; summarize.
- Be decisive and brief; do NOT provide step-by-step reasoning chains.
- Do NOT try to reason about and answer the question yourself. The question is only to provide context for evaluating the PREDICTION against the GROUND_TRUTHS.

\end{promptbox}

\begin{promptbox}[Data Cleaning Prompt]
You are an expert QA dataset evaluator.

Given a question and answers separated by <|answer_split|>, determine if this data is high-quality based on these criteria:
1. The question is a clear, complete question (not a statement or fragment)
2. All provided answers correctly answer the question
3. The answers are the only correct answers (no other valid answers exist)

First, explain your reason.
Then, respond with exactly one word: "Yes", "No", or "Unsure", wrapped in tags <quality></quality>.

Question: {question}
Answers: {answers}
\end{promptbox}

\end{document}